\documentclass[journal]{EMO2007LBP}
\usepackage{algorithm}
\usepackage{algorithmic}
\usepackage{graphicx}

\newtheorem{definition}{Definition}[section]
\usepackage{url}       % url.sty is already installed on most LaTeX
\usepackage{amsmath}   % From the American Mathematical Society
\begin{document}

%
% paper title
\title{Bin Packing Under Multiple Objectives --\\a Heuristic Approximation Approach}

% author names
% note positions of commas and nonbreaking spaces ( ~ ) LaTeX will not break
% a structure at a ~ so this keeps an author's name from being broken across
% two lines.
% use \thanks{} to gain access to the first footnote area
% a separate \thanks must be used for each paragraph as LaTeX2e's \thanks
% was not built to handle multiple paragraphs
\author{Martin Josef Geiger%
\thanks{Martin Josef Geiger is with the Department of Industrial
Management (510A), University of Hohenheim, 70593 Stuttgart, Germany
(phone: 0049-711-45923462; fax: 0049-711-45923232; email:
mjgeiger@uni-hohenheim.de).}}
% note the % following the last name and also the first \thanks -
% these prevent an unwanted space from occurring between the last author name
% and the end of the author line. i.e., if you had this:
%
% \author{....lastname \thanks{...} \thanks{...} }
%                     ^------------^------------^----Do not want these spaces!
%
% a space would be appended to the last name and could cause every name on that
% line to be shifted left slightly. This is one of those "LaTeX things". For
% instance, "A\textbf{} \textbf{}B" will typeset as "A B" not "AB". If you want
% "AB" then you have to do: "A\textbf{}\textbf{}B"
% \thanks is no different in this regard, so shield the last } of each \thanks
% that ends a line with a % and do not let a space in before the next \thanks.
% Spaces after \IEEEmembership other than the last one are OK (and needed) as
% you are supposed to have spaces between the names. For what it is worth,
% this is a minor point as most people would not even notice if the said evil
% space somehow managed to creep in.
%

% *** Note that you probably will NOT want to include the author's name in ***
% *** the headers of peer review papers.                                   ***

% If you want to put a publisher's ID mark on the page
% (can leave text blank if you just want to see how the
% text height on the first page will be reduced by IEEE)
%\pubid{0000--0000/00\$00.00~\copyright~2002 IEEE}

% use only for invited papers
%\specialpapernotice{(Invited Paper)}

% make the title area

\maketitle
\thispagestyle{empty}

\begin{abstract}
The article proposes a heuristic approximation approach to the bin
packing problem under multiple objectives. In addition to the
traditional objective of minimizing the number of bins, the
heterogeneousness of the elements in each bin is minimized, leading
to a biobjective formulation of the problem with a tradeoff between
the number of bins and their heterogeneousness. An extension of the
Best-Fit approximation algorithm is presented to solve the problem.
Experimental investigations have been carried out on benchmark
instances of different size, ranging from 100 to 1000 items.
Encouraging results have been obtained, showing the applicability of
the heuristic approach to the described problem.

\end{abstract}

% Note that keywords are not normally used for peerreview papers.

% For peer review papers, you can put extra information on the cover
% page as needed:
% \begin{center} \bfseries EDICS Category: 3-BBND \end{center}
%

%=================================================================
%-----------------------------------------------------------------
% INTRODUCTION
%-----------------------------------------------------------------
%=================================================================
\section{\label{sec:introduction}Introduction}
\PARstart{T}{HE} term ``bin packing'' describes a class of
well-known, classical problems with numerous applications in
logistics, operations research and related disciplines. From single
dimensional to multi-dimensional problems, various types can be
identified in practice. Common to all is the overall task of packing
a finite number of $n$ items into a minimum number of bins
(knapsacks) subject to a set of practical constraints and
requirements. These include given capacities of the bins, but also
other considerations such as irregularly shaped bins, load balancing
of the bins, etc.

Numerous approaches including exact, heuristic, and metaheuristic
algorithms have been proposed for the resolution of bin packing
problems, and a rich literature on packing problems exists, with
important classifications by {\sc Dyckhoff}
\cite{dyckhoff:1990:article} and more recently {\sc W\"{a}scher} et
al. \cite{wascher:2006:techreport}. While the majority of approaches
is dedicated to single-objective models, only minimizing the number
of bins used, the multi-objective nature of many of these problems
becomes more and more obvious. Following early work of {\sc
W\"{a}scher} \cite{wascher:1990:article}, modern heuristics such as
Particle Swarm Optimization have recently been applied to a
multi-objective variant of the two-dimensional bin-packing problem
\cite{liu:2006:inproceedings}. For the here considered
multi-objective bin-packing problem however, no corresponding
studies have been carried out to our knowledge.

The remainder of the article is organized as follows.
Section~\ref{sec:problem:formulation} describes the multi-objective
bin packing problem and its' underlying practical application. A
heuristic approximation approach is presented in the following
Section~\ref{sec:heuristic}. In brief, we propose an extension of
the well-known best-fit heuristic, allowing the computation of a set
of solutions that constitute an approximation to the set of
efficient solutions. Experimental results of the approach to the
problem are reported in Section~\ref{sec:experiments}, and
conclusions are given in Section~\ref{sec:conclusions}.

\section{\label{sec:problem:formulation}Problem description}
We consider a bin packing problem where a given number of $n$ items
has to be packed into $n$ bins, each of capacity $c$
\cite{martello:1990:book}. Each item $j$ is characterized by a
weight $w_{j}$ and an additional attribute $a_{j}$. While the
weights refer to the size of the items and therefore have to be
taken into consideration when filling up a bin to at most its'
capacity $c$, the attributes $a_{j}$ describe properties of the
items on a nominal scale. On the basis of this description, a
comparison of two items $i,j$ is possible such that they are either
identical with respect to $a_{i}$ and $a_{j}$, $a_{i} = a_{j}$ or
not: $a_{i} \neq a_{j}$. The goal of packing the items into bins can
then be modeled as follows.

\begin{eqnarray}
\label{eqn:z1}\mbox{minimize} \quad z_{1} = \sum_{i=1}^{n} y_{i}\\%
\label{eqn:z2}\mbox{minimize} \quad z_{2} = \frac{1}{z_{1}}\sum_{i=1}^{n} u_{i}\\%
\label{eqn:max:capacity}\mbox{s.\,t.} \quad \sum_{j=1}^{n} w_{j} x_{ij} \leq c y_{i} && i\in N = \{1,\ldots,n\},\\%
\label{eqn:assign:all}\sum_{i=1}^{n}x_{ij}  = 1 && j\in N,\\%
y_{i} = 0\;\, \mbox{or}\;\, 1 && i \in N,\\%
x_{ij} = 0\;\, \mbox{or}\;\, 1 && i \in N, j \in N,%
\end{eqnarray}

where

\begin{eqnarray*}
y_{i} & = & \left\{ \begin{array}{rl}1 & \mbox{if bin $i$ is used}\\
0 & \mbox{otherwise}
\end{array}\right.\\
x_{ij} & = & \left\{ \begin{array}{rl}1 & \mbox{if item $j$ is assigned to bin $i$}\\
0 & \mbox{otherwise}
\end{array}\right.\\
u_{i} && \mbox{counts the number of distinct attributes in bin $i$}
\end{eqnarray*}

Expression~(\ref{eqn:z1}) minimizes the number of bins. The second
objective given in~(\ref{eqn:z2}) minimizes the average
heterogeneousness of the bins. To do this, the number of distinct
attributes $u_{i}$ is counted for each bin $i$. Unused bins ($y_{i}
= 0$) have a value of $u_{i} = 0$. Used bins ($y_{i} = 1$) have a
possible minimum value of $u_{i} = 1$. This is the case when all
items in the particular bin have the identical nominal attribute.
The values of $u_{i}$ are bounded by either the number of items
assigned to a bin or the number of distinct attributes over all
items $i$.

Intuitively, the two objective functions are of conflicting nature.
While a large number of bins allows the packing of bins which are
each fully homogeneous, leading to a $z_{2} = 1$, a solution being
minimal for $z_{1}$ will require the packing of items $i,j$ of
different $a_{i}$ and $a_{j}$ into the same bin. It can therefore be
suspected that not a single solution $x$ exists in the set of
feasible solutions $X$ that equally minimizes both objective
functions $z_{1}$ and $z_{2}$. In brief, this leads to a vector
optimization problem in which a solution $x \in X$ is evaluated with
respect to a vector $Z(x) = \left( z_{1}(x), z_{2}(x) \right)$. The
resolution of the problem has consequently to be seen in the
identification of all efficient outcomes or the Pareto set $P$,
introduced in the following Definitions~\ref{def:dominance} and
\ref{def:efficiency}.

\begin{definition}[Dominance]\label{def:dominance}
A vector $Z(x)$ is said to dominate $Z(x')$ iff $z_{k}(x) \leq
z_{k}(x') \forall k=1, \ldots, K \wedge \exists k \mid z_{k}(x) <
z_{k}(x')$. We denote the dominance of $Z(x)$ over $Z(x')$ with
$Z(x) \preceq Z(x')$.
\end{definition}

\begin{definition}[Efficiency,
Pareto-optimality]\label{def:efficiency} The vector $Z(x), x \in X$
is said to be efficient iff $\neg\exists Z(x'), x' \in X \mid Z(x')
\preceq Z(x)$. The corresponding alternative $x$ is called {\em
Pareto-optimal}, the set of all Pareto-optimal alternatives {\em
Pareto-set $P$}.
\end{definition}

Numerous practical applications of the formal model exist. In many
cases, a minimum number of bins should be used when packing a given
set of items, however assuring a maximum possible homogeneousness of
the items being packed together into a single bin. Applications
include the storage of goods, the storage of music/video data on
optical discs, etc.

\section{\label{sec:heuristic}A heuristic approximation approach}
As already mentioned in Section~\ref{sec:introduction}, numerous
algorithms have been proposed to solve the single-objective variant
of the bin packing problem. One important heuristic is the {\em
Best-Fit} algorithm, important for both for its' time complexity as
well as for its' worst-case complexity
\cite{coffmann:1984:incollection}. Best-Fit subsequently assigns
items to the feasible bin having the smallest residual capacity. If
no such bin exists, the item is assigned to a previously unused
(new) bin.

Unfortunately however, Best-Fit only takes into consideration the
weights $w_{j}$ of the items and the residual capacities of the bins
when selecting the `best-fitting'-bin. In order to address the
problem described in Section~\ref{sec:problem:formulation}, a method
of controlling the heterogeneousness of the bins needs to be
included in the algorithm. Algorithm~\ref{alg:Best:Fit} describes
such an attempt, allowing the successive computation of alternatives
with different heterogeneousness levels and therefore providing an
idea of how to compute an approximation to the vector optimization
problem given in Section~\ref{sec:problem:formulation}.

\begin{algorithm}[!ht]%
\caption{\label{alg:Best:Fit}Multi-objective Best-Fit algorithm}
\begin{algorithmic}[1]%
\REQUIRE{$s$, $\overline{m}$}%
\STATE{Compute the maximum possible heterogeneousness of a bin $\overline{u}$}%
\STATE{Set $u = 1$}%
\STATE{$P^{approx} = \emptyset$}%
\REPEAT%
    \FOR{$m=1$ to $\overline{m}$}
        \STATE{Construct a new solution $x$:}
        \FORALL{$n$ items}%
            \STATE{Compute the maximally allowed heterogeneousness $u_{max}$ of the Best-Fit-bin:\\
            $u_{max} = \lfloor u \rfloor$ with probability $1 - \left( u - \lfloor u \rfloor \right)$
            and $u_{max} = \lceil u \rceil$ with probability $ u - \lfloor u \rfloor$}%
            \STATE{Compute the Best-Fit-bin with respect to the randomly determined $u_{max}$}%
            \STATE{Assign $i$ to the Best-Fit-bin}%
        \ENDFOR
        \STATE{Update $P^{approx}$ with $x$:\\
        Remove all elements in $P^{approx}$ which are dominated by $Z(x)$;\\
        Add $x$ to $P^{approx}$ if $Z(x)$ is not dominated by any element in $P^{approx}$}%
    \ENDFOR
    \STATE{Set $u = u + s$}%
\UNTIL{$u > \overline{u}$}%
\STATE{Return $P^{approx}$}%
\end{algorithmic}%
\end{algorithm}

The modified Best-Fit algorithm is based in principle on the
conventional method. However, in order to control the
heterogeneousness of the bins, an additional control parameter
$u_{max}$ is used as described in step 8 of
Algorithm~\ref{alg:Best:Fit}. Starting with an initial value of
$u_{max} = u = 1$, only Best-Fit-bins are allowed which are fully
homogeneous. This means that an item may only be assigned to a bin
containing other elements of identical attributes $a_{j}$. In this
stage of the algorithm, solutions are computed that lead to the
lowest possible value of $z_{2} = 1$ as all bins contain homogeneous
items.

With increasing value of $u$, incremented by $s$ in step 14 of the
algorithm, Best-Fit-bins become possible that have a higher
heterogeneousness. This concept is randomized throughout the
generation of the solutions, allowing a gradual transition from
$u_{max} = u = 1$ to the maximum possible heterogeneousness $u_{max}
= \overline{u}$. Due to the randomness in the algorithm, different
runs lead to different outcomes. We therefore propose to compute a
number of solutions with each setting of $u$, given as control
parameter $\overline{m}$.

Throughout the algorithm, an archive $P^{approx}$ of the best
solutions is kept which is returned after the algorithm terminates.
This archive represents an approximation to the true Pareto-set $P$.

\section{\label{sec:experiments}Experimental investigation}
\subsection{Generation of test instances and experimental setup}
In order to test the effectiveness of the multi-objective extension
of the Best-Fit approximation algorithm, four multi-objective test
instances have been computed with values of $n = 100$, $n = 200$, $n
= 500$, and $n = 1000$. The data has been derived taking
$\frac{n}{5}$ bins, each of capacity $c = 1000$, and randomly
splitting the capacity into five items $j, \ldots, j+4$ such that
the weights $w_{j}$ of the items add up to $c$: $\sum_{j}^{j+4}
w_{j} = c$. This means that the so constructed instances have a
solution for which the minimum number of bins is $\frac{n}{5}$ and
therefore equal to the trivial lower bound $\left\lceil
\frac{\sum_{j=1}^{n} w_{j}}{c} \right\rceil$.

The items of each instance have been randomly assigned nominal
attributes from a set of five different attributes. Each attribute
has been selected with equal probability of $0.2$.

We tested the randomized Best-Fit algorithm from
Section~\ref{sec:heuristic} on the proposed benchmark instances with
control parameters $s = 0.1$ and $\overline{m} = 100$. This means
that the probability of selecting a Best-Fit-bin of maximum
heterogeneousness $u_{max} = \lceil u \rceil$ over a Best-Fit-bin of
maximum heterogeneousness $u_{max} = \lfloor u \rfloor$ increases in
each step 14 of Algorithm~\ref{alg:Best:Fit} with 10\%. The inner
loop of the algorithm computes $\overline{m} = 100$ solutions with
each setting of the parameters.

In addition to the multi-objective Best-Fit algorithm, a Random-Fit
algorithm has been tested for comparison reasons. This algorithm
randomly selects a bin that allows the assignment of the currently
considered item with respect to the chosen maximum heterogeneousness
$u_{max}$. Apart from that aspect, the algorithm is implemented
identically to the pseudo-code given in
Algorithm~\ref{alg:Best:Fit}.

Three different input sequences of the items have been tested:
\begin{itemize}
\item Sorting the items in decreasing order of $w_{j}$
\item Sorting the items in increasing order of $w_{j}$
\item Arranging the items in random order
\end{itemize}

\subsection{Results}
The experimental investigations revealed that only few efficient
outcomes exist for the instances. Instead of plotting the outcomes
in figures, we chose to give the data of all found best vectors
$Z(x) = \left( z_{1} (x), z_{2} (x) \right)$. The following
Table~\ref{tbl:results:100} shows the results for the smallest
instance with $n = 100$. It can be seen, that both Best-Fit and
Random-Fit perform comparably good given a decreasing or random
order of the items.

\begin{table}[!ht]
\caption{\label{tbl:results:100}Identified vectors $(z_{1}, z_{2})$
for the instance with $n = 100$. The best outcomes are highlighted
bold.} \centering
\begin{tabular}{lcc}
\hline%%
Item order & Best-Fit & Random-Fit\\%%
\hline%%
Decreasing $w_{j}$ & \textbf{(22, 1.000)} & \textbf{(22, 1.000)}\\%%
           & (21, 2.048) & (21, 1.952)\\%%
\hline%%
Increasing $w_{j}$ & (25, 1.000) & (25, 1.000)\\%%
           & (24, 1.108) & (24, 1.125)\\%%
\hline%%
Random     & \textbf{(22, 1.000)} & \textbf{(22, 1.000)}\\%%
           & \textbf{(21, 1.190)} & (21, 1.952)\\%%
\hline%%
\end{tabular}
\end{table}

Similar results have been obtained for the instance with $n = 200$
as shown in Table~\ref{tbl:results:200}. Again, Best-Fit and
Random-Fit lead to the best results given an order of the items with
decreasing $w_{j}$. While both identify the vector (43, 1.000), they
are incomparable with respect to the other best found outcomes (42,
1.214) and (41, 1.902).

\begin{table}[!ht]
\caption{\label{tbl:results:200}Identified vectors $(z_{1}, z_{2})$
for the instance with $n = 200$. The best outcomes are highlighted
bold.} \centering
\begin{tabular}{lcc}
\hline%%
Item order & Best-Fit & Random-Fit\\%%
\hline%%
Decreasing $w_{j}$ & \textbf{(43, 1.000)} & \textbf{(43, 1.000)}\\%%
           & \textbf{(42, 1.214)} & (42, 1.881)\\%%
           & (41, 1.927) & \textbf{(41, 1.902)}\\%%
\hline%%
Increasing $w_{j}$ & (50, 1.000) & (50, 1.000)\\%%
           & (49, 1.163) & (49, 1.531)\\%%
           & (48, 1.563)\\%%
\hline%%
Random     & (43, 1.000) & (43, 1.000)\\%%
           & (42, 1.381) & (42, 1.929)\\%%
           & (41, 1.927) & (41, 2.000)\\%
\hline%%
\end{tabular}
\end{table}

For the next instance with $n = 500$, given in
Table~\ref{tbl:results:500}, Random-Fit appears to lead to superior
results, however with a very small distance to Best-Fit. While the
assignment of items in increasing order of $w_{j}$ is clearly
inferior, the random ordering of items appears to become less and
less favorable in comparison to the decreasing order.

\begin{table}[!ht]
\caption{\label{tbl:results:500}Identified vectors $(z_{1}, z_{2})$
for the instance with $n = 500$. The best outcomes are highlighted
bold.} \centering
\begin{tabular}{lcc}
\hline%%
Item order & Best-Fit & Random-Fit\\%%
\hline%%
Decreasing $w_{j}$ & \textbf{(102, 1.000)} & \textbf{(102,
1.000)}\\%%
           & (101, 2.020) & \textbf{(101, 1.911)}\\%%
\hline%%
Increasing $w_{j}$ & (127, 1.000) & (127, 1.000)\\%%
           &              & (126, 1.135)\\%%
           & (125, 1.136) & (125, 1.152)\\%%
           & (124, 1.177)\\%%
           & (123, 1.325)\\%%

\hline%%
Random     &              & (104, 1.000)\\%%
           &              & (103, 1.951)\\%%
           & \textbf{(102, 1.000)} & (102, 1.971)\\%%
           & (101, 1.972) & (101, 3.168)\\%%
\hline%%
\end{tabular}
\end{table}

The results of the largest instance, shown in
Table~\ref{tbl:results:1000}, confirm that the ordering of the items
becomes more influential with increasing size of the instance. The
best results have been obtained assigning the items in decreasing
order of $w_{j}$ while the random ordering turned out to be
comparably weak. When comparing Best-Fit and Random-Fit, Best-Fit
appears to lead to more efficient outcomes, but still Random-Fit is
able to identify solutions that have not been found by the Best-Fit
algorithm.

\begin{table}[!ht]
\caption{\label{tbl:results:1000}Identified vectors $(z_{1}, z_{2})$
for the instance with $n = 1000$. The best outcomes are highlighted
bold.} \centering
\begin{tabular}{lcc}
\hline%%
Item order & Best-Fit & Random-Fit\\%%
\hline%%
Decreasing $w_{j}$ &              & (205, 1.000)\\%%
           & \textbf{(203, 1.000)}\\%%
           & \textbf{(202, 1.287)} & (202, 1.906)\\%%
           &              & \textbf{(201, 1.910)}\\%%
\hline%%
Increasing $w_{j}$ & (250, 1.000) & (250, 1.000)\\%%
           & (249, 1.177) & (249, 1.639)\\%%
           & (248, 1.194)\\%%
\hline%%
Random     & (205, 1.000) & (205, 1.000)\\%%
           & (204, 1.140) & (204, 1.971)\\%%
           & (203, 1.897) & (203, 2.739)\\%%
           & (202, 1.965) & (202, 3.238))\\%%
           & (201, 2.781)\\%%
\hline%%
\end{tabular}
\end{table}

Common to the results of all investigated instances is that the
approximation algorithms have not been able to identify the minimal
solution for $z_{1}$. The best solutions with respect to $z_{1}$ are
still one bin larger than the minimum possible value.

\section{\label{sec:conclusions}Conclusions}
The article presented a study on the multi-objective bin packing
problem. We considered the objective of minimizing the number of
bins as well as the objective of minimizing the average
heterogeneousness of the bins, based on nominal attributes of the
items. The two conflicting objectives led to the formulation of the
problems as a vector optimization problem.

A modified Best-Fit approximation algorithm has been presented to
compute an approximation of the set of efficient solutions. The
procedure allows the controlled consideration of the
heterogeneousness of the bins, integrating the parameter in the
selection process of the bins in a randomized fashion.

Experimental investigations have been carried out on a set of
benchmark instances, and comparison results have been obtained from
a Random-Fit heuristic. The results are encouraging, as very close
approximations to the set of efficient outcomes have been
identified. It has become clear, that with growing size of the
instances, measured by the number of items $n$, the processing order
of the items plays an increasingly important role. For small
instances, a random order of the items turns out to be feasible. For
large instances however, an order of decreasing $w_{j}$ is necessary
to obtain good results.

In conclusion, the presented multi-objective Best-Fit algorithm led
to satisfying solutions. The running times of the approach remained
on an Intel Pentium IV 1.8 GHz processor within a few seconds for
each test run. We conclude that the algorithm may also be beneficial
when computing a first, qualitatively good approximation of the
Pareto-set which is then used in a more complex improvement
(meta-)heuristic.

\newpage

\section*{Acknowledgment}
The participation in the 4th International Conference on
Evolutionary Multi-Criterion Optimization (EMO2007) has been
partially supported by the Deutsche Forschungsgemeinschaft (DFG),
grant no.\ 535530.

\bibliography{../../../lit_bank,../../../lit_bank_nv,../../../lit_bank_datei}
\bibliographystyle{IEEEtran.bst}

\end{document}